\documentclass{article}

\usepackage[final,nonatbib]{neurips_2020}  

\usepackage[utf8]{inputenc} 
\usepackage[T1]{fontenc}    
\usepackage{hyperref}       
\usepackage{url}            
\usepackage{booktabs}       
\usepackage{amsfonts}       
\usepackage{nicefrac}       
\usepackage{microtype}      
\usepackage[square,comma,numbers,sort&compress]{natbib} 
\usepackage{amsmath, amssymb}
\usepackage{graphicx}
\graphicspath{./figures}

\newcommand{\methodname}{embedded variance}
\newcommand{\comparename}{learned variance}

\title{A Framework for Variational Inference of Lightweight Bayesian Neural
    Networks with Heteroscedastic Uncertainties}

\author{%
    David J. Schodt \\ Teledyne Scientific \& Imaging \\
    \And
    Ryan Brown \\ Teledyne Scientific \& Imaging \\
    \And
    Michael Merritt \\ Teledyne Scientific \& Imaging \\
    \And
    Samuel Park \\ Teledyne Scientific \& Imaging \\
    \And
    Delsin Menolascino \\ Teledyne Scientific \& Imaging \\
    \And
    Mark A. Peot \\ Teledyne Scientific \& Imaging \\
}

\begin{document}

\maketitle

\begin{abstract}
Obtaining heteroscedastic predictive uncertainties from a Bayesian
Neural Network (BNN) is vital to many applications.  Often, heteroscedastic
aleatoric uncertainties are learned as outputs of the BNN in addition to the
predictive means, however doing so may necessitate adding more learnable
parameters to the network.  In
this work, we demonstrate that both the heteroscedastic aleatoric and
epistemic variance can be embedded into the variances of learned BNN
parameters, improving predictive performance for lightweight networks.
By complementing this approach with a moment propagation approach to
inference, we introduce a relatively simple framework for
sampling-free variational inference suitable for lightweight BNNs.
\end{abstract}

\section{Introduction}
Neural networks are a broad set of models capable of learning complex data
trends.  If we feed a (sufficiently complex) trained neural network data $x_i$
sampled from the same distribution as the training data, we expect the output
$\hat{y}(x_i)$ to be similar to the ground truth $y(x_i)$ in terms of a chosen
metric (e.g., in regression, mean squared error is often used to define
similarity between predictions and the ground truth observations).  In other
words, we expect the
neural network output to fall within the range of values expected given the
uncertainties inherent to the problem.  Unfortunately, classical neural
networks do
not provide a natural mechanism for prescribing uncertainty to their outputs.
For some applications, this may not pose a problem, as we can enforce
restrictions on the type of data fed to the network.  For many applications,
however, this shortcoming prevents or severely discourages the use of neural
networks.

Bayesian Neural Networks (BNNs) \cite{mackay_practical_1992,
    neal_bayesian_1996} extend the capabilities of classical neural
networks by representing their parameters and/or outputs as statistical
distributions
with non-zero variance.  By training a BNN to minimize an appropriate loss
criterion, the parameter distributions can be guided to represent
meaningful predictive uncertainties in the network outputs.  BNNs thus
enable us to quantify their output uncertainty in a principled manner, making
BNNs
attractive for applications requiring competency assessments of their
performance.

Broadly speaking, we wish to account for two types of uncertainties in the
outputs of a BNN \cite{kendall_what_2017}: (1) epistemic uncertainties, or
those arising from the model itself, and (2) aleatoric uncertainties, or
those inherent to the data generation process.  Epistemic uncertainty can
be used, for example, to identify the model's uncertainty in its own
prediction, such as that for an input which was dissimilar to data in the
training distribution.  On the other hand, aleatoric uncertainty
guides our expectations for the output predictions; we can only
improve the precision of our prediction within the limits allowed
by noisy data.  For many applications, we
additionally desire \emph{heteroscedastic} uncertainties, or uncertainties
which depend on specific inputs.  For example, some data might be noisier
than others, and the predictive aleatoric uncertainty should reflect this.

In the context of variational inference of BNNs, the epistemic uncertainty
is often extracted from the variance of BNN weights and biases
\cite{gal_uncertainty_2016, gal_dropout_2016,
    kendall_what_2017, wu_deterministic_2019,
    postels_sampling-free_2019, hausmann_sampling-free_2020}.
As we propagate an input through the BNN, the resulting output accumulates the
uncertainties of these weights and biases.  This
accumulated uncertainty in the BNN prediction is then interpreted as the
epistemic uncertainty of the model.  The aleatoric uncertainty is then
incorporated as either a homoscedastic uncertainty
\cite{gal_dropout_2016, gal_uncertainty_2016, hausmann_sampling-free_2020}
or is learned as an additional
(heteroscedastic) output of the network \cite{nix_1994, an_variational_2015,
    kendall_what_2017,
    wu_deterministic_2019, antoran_getting_2021}.

We propose a straightforward reinterpretation of BNN
inference which provides aleatoric and epistemic
predictive uncertainties in a heteroscedastic manner.  Instead of
distinguishing the aleatoric and epistemic uncertainties, we seek to
characterize the combined total uncertainty during BNN inference, allowing
us to embed both the aleatoric and epistemic contributions directly in the
variances of BNN parameters.  We demonstrate that our revised inference
approach can improve the predictive performance of BNNs by eliminating the need
for additional output nodes to predict aleatoric uncertainties, an improvement
which is especially useful when
designing lightweight BNNs (e.g., BNNs intended for deployment on systems
with hardware limitations or compute time constraints).
We further combine
our revised approach to inference with a moment propagation scheme
similar to those used in Refs. \cite{wu_deterministic_2019,
    hausmann_sampling-free_2020}, providing a lightweight framework for
variational inference of BNNs with heteroscedastic predictive uncertainties.

\section{Background and motivation}
\label{sec:background}
\subsection{Bayesian neural networks}
\label{sec:bnn_intro}
In this section, we briefly introduce the notation used
throughout this work.  For detailed reviews of BNNs, we refer readers
to Refs. \cite{shridhar_comprehensive_2019, gawlikowski_survey_2022,
    jospin_hands-bayesian_2022}.

We consider BNNs to be neural networks
whose outputs can be used to define predictive distributions, where some
prior information is used to guide the distributions of network parameters
(or some subset thereof).  For a typical feedfoward
BNN denoted by $f(\cdot | \mathbf{w})$, we use the notation $\hat{y} =
f(x | \mathbf{w})$ where $\hat{y}$ represents the output of
the BNN given an input $x$, and $\mathbf{w}$ represents the internal
parameters of the BNN.  We denote a prior distribution over the BNN
parameters as $p(\mathbf{w})$.  The variational posterior over the
$M$ learnable BNN parameters is taken
to be a product of normal distributions as $q(\mathbf{w} |
\theta) = \prod_{j=1}^M q(w_j | \mu_j, \sigma_j^2) = \prod_{j=1}^M
\mathcal{N}(w_j |
\mu_j,
\sigma_j^2)$ where $\theta = \{(\mu_j, \sigma_j^2)\}_{j=1}^M$ are the
parameters of normal distributions with mean $\mu_j$ and variance
$\sigma_j^2$.  The set of $N$
input/output pairs in the training data is denoted as $\mathcal{D} \equiv
\{(x_k, y_k)\}_{k=1}^N$.

\subsection{Evidence lower bound}
Under the variational approach, BNNs are often trained to minimize the
negative of the Evidence Lower
Bound (ELBO), which is the Kullback-Leibler (KL) divergence
between the true and the variational posterior distributions over the BNN
parameters \cite{graves_practical_2011, kingma_auto-encoding_2013,
    blundell_weight_2015,
    shridhar_comprehensive_2019, jospin_hands-bayesian_2022}.  The negative
ELBO between the true and variational posteriors $p(\mathbf{w} |
\mathcal{D})$ and $q(\mathbf{w} | \theta)$, respectively, can be written
as
\begin{equation*}
    \mathcal{L}(\mathcal{D}, \theta) = \mathrm{KL}[q(\mathbf{w} | \theta) || p(\mathbf{w})]
    - E_{q(\mathbf{w} | \theta)} [\log p(\mathcal{D} | \mathbf{w})]
\end{equation*}
where $\mathbf{w}$ are the network parameters, $\theta$ are the parameters
of
the variational posterior distribution, and $\mathcal{D}$ is the data.  An
additional scaling factor $\alpha$ is sometimes used to modulate the impact
of the prior distribution, giving
\begin{equation}
    \label{eqn:elbo_loss}
    \mathcal{L}(\mathcal{D}, \theta)
    = \alpha \mathrm{KL}[q(\mathbf{w} | \theta) || p(\mathbf{w})]
    - E_{q(\mathbf{w} | \theta)} [\log p(\mathcal{D} | \mathbf{w})]
\end{equation}
The first term in
Eqn. \ref{eqn:elbo_loss} penalizes deviations of the posterior from the
prior (thus acting as a regularizer), while the reconstruction loss
$-E_{q(\mathbf{w} | \theta)}
[\log p(\mathcal{D} | \mathbf{w})]$ penalizes deviations of the predictive
distributions from the data.

\subsubsection{Computing the Kullback-Leibler divergence}
The KL divergence $\mathrm{KL}[q(\mathbf{w} | \theta) || p(\mathbf{w})]$
can be computed analytically for some
commonly used prior-posterior pairs, such as when the prior and
the variational posterior are both multivariate normal distributions
\cite{wu_deterministic_2019}.  Alternatively, when a simple closed-form
expression for $\mathrm{KL}[q(\mathbf{w} | \theta) || p(\mathbf{w})]$ is
not available, this term can be approximated from $N$ Monte Carlo samples
as \cite{kingma_auto-encoding_2013, blundell_weight_2015}
\begin{equation}
    \label{eqn:kl_monte_carlo}
    \mathrm{KL}[q(\mathbf{w} | \theta) || p(\mathbf{w})] \approx
    \frac{1}{N} \sum_{i=1}^N \left[ \log{q(\mathbf{w}^{(i)} | \theta)} -
    \log{p(\mathbf{w}^{(i)})} \right]
\end{equation}
where $w^{(i)}$ is a single draw of weights from the variational
posterior.  In our experience, a single Monte Carlo sample ($N=1$) of
$\mathrm{KL}[q(\mathbf{w} | \theta) || p(\mathbf{w})]$ per
minibatch is often sufficient to promote regularization.

\subsubsection{Computing the reconstruction loss}
Computing the reconstruction loss
$-\mathbb{E}_{q(\mathbf{w} | \theta)} [\log p(\mathcal{D} |
\mathbf{w})]$ can be difficult in many applications.  For arbitrarily
complex BNNs, the
distributional form of $p(\mathcal{D} | \mathbf{w})$ is unavailable, even
for the relatively simple multivariate normal variational posterior used in
this work.
For regression, this forces us to either restrict the BNN architecture to
permit a closed-form $\mathbb{E}_{q(\mathbf{w} | \theta)} [\log
p(\mathcal{D} | \mathbf{w})]$ (e.g., by only allowing the output layer to
contain Bayesian neurons) or to approximate the reconstruction term (e.g.,
by assuming normality and moment matching the corresponding mean and
variance).

\subsection{Computing uncertainties with Bayesian neural networks}
\label{sec:bnn_uncertainty_bg}
By allowing their weights and biases to take on distributions with non-zero
variance, BNNs enable expression of predictive uncertainty in their outputs.
Refs. \cite{gal_uncertainty_2016, kendall_what_2017,
    gawlikowski_survey_2022}
provide discussions about the different types of uncertainties that we wish
to extract from
BNN, namely aleatoric and epistemic uncertainty.  Each of these uncertainties
can be
either homoscedastic (constant for all $x$) or heteroscedastic
(varying with $x$).  As observed in \cite{gal_uncertainty_2016,
    kendall_what_2017}, much of the research involving BNN uncertainty focuses
on expressing either aleatoric or epistemic uncertainty, but not both.
Furthermore, aleatoric uncertainty is often modeled to be homoscedastic
\cite{gal_uncertainty_2016, springenberg_bayesian_2016,
    hausmann_sampling-free_2020}, a simplification which is too restrictive for
many applications.

Focusing our attention to regression, we assume in this work that each
input output pair $\{x_k, y_k\}$ in the data is normally distributed as
\begin{equation}
    \label{eqn:normal_data_assumption}
    p(\mathcal{D}_k) = p(y_k | x_k) = \mathcal{N}(\mu_{y,k}, \sigma_{a,k}^2)
\end{equation}
where $\mu_{y,k}$ is the mean and $\sigma_{a,k}^2$ is the (heteroscedastic)
aleatoric
variance, hence estimating aleatoric uncertainty in the network's prediction
involves estimating the variance of this distribution.

\subsubsection{Learning the aleatoric uncertainty as a network output}
\label{sec:learning_mean_var}
One method to obtain the aleatoric variance $\sigma_{a,k}^2$ is to
learn it directly as an output of the network \cite{nix_1994,
an_variational_2015,
    kendall_what_2017, wu_deterministic_2019, antoran_getting_2021}.  Using the
notation defined
in Section \ref{sec:bnn_intro}, we write this as $\hat{y} = f(x |
\mathbf{w}) = [\mu_{y}, \sigma_a^2]$ to emphasize that the BNN
output consists of both a predictive mean and a corresponding aleatoric
variance.

Although learning the variance as an additional output of the BNN is a
simple and effective approach to retrieving aleatoric uncertainty, it can
require us to choose a bulkier
network than needed to just predict the mean $\mu$.  In
other words, a lightweight BNN capable of learning a single output may not
be sufficiently expressive to predict the variance as an additional output,
requiring us to add more learnable parameters to the BNN to achieve our
desired performance.

\section{Proposed inference framework}
As discussed in Section \ref{sec:learning_mean_var}, the aleatoric variance
$\sigma_y^2$ can be
predicted as an additional output of the BNN, however requiring this
additional output node may be undesirable when seeking lightweight models.
To eliminate this burden, we instead propose embedding the
aleatoric uncertainty into the distributions of the network parameters.  In
this manner, we obtain the \emph{total} uncertainty in the BNN's prediction
without splitting the network head to produce additional
outputs or sacrificing heteroscedasticity.  We additionally describe a
simplified moment propagation strategy for inference similar to those
presented in Refs. \cite{wu_deterministic_2019,
    hausmann_sampling-free_2020}, providing a framework for lightweight,
computationally efficient variational inference of BNNs with
heteroscedastic aleatoric and epistemic uncertainty.  We describe each of
these components of our framework in the following sections.

\subsection{Embedding total uncertainty into Bayesian neural network
    parameters}
The standard approach to computing epistemic uncertainty from a BNN builds
upon the interpretation that the distributions of the Bayesian neurons
represent the epistemic uncertainty.  Given more training
data, the variances of individual Bayesian neurons will presumably shrink,
indicating that the BNN is more certain about its predictions.  In this
interpretation, the epistemic uncertainty directly corresponds to
variance in the outputs of the BNN, and hence can be computed from repeated
Monte Carlo samples of the BNN \cite{gal_uncertainty_2016, gal_dropout_2016,
    kendall_what_2017} or deterministically \cite{wu_deterministic_2019,
    postels_sampling-free_2019, hausmann_sampling-free_2020}.  In other words,
the variance of the BNN output is attributable only to the epistemic
uncertainty, requiring us to predict the aleatoric uncertainty as an
additional BNN output.

Many applications requiring uncertainty in BNN predictions do not require
us to distinguish between epistemic and aleatoric uncertainties; as long as
the predicted uncertainty is the \emph{total} uncertainty, the
distinction of underlying components is inconsequential (see, however, Ref.
\cite{antoran_getting_2021} for an application requiring this
distinction).  For example, a BNN might be deployed in a high-consequence
system capable of contacting a human operator, in which case highly
uncertain predictions should elicit human intervention regardless of their
cause.  As such, we opt to combine the two categories of uncertainty,
allowing us to train the Bayesian neurons to intrinsically reproduce the
combined epistemic and aleatoric uncertainties.  This corresponds to a
reinterpretation of the distributions of the Bayesian neurons: instead of
encoding only the model uncertainty, their distributions encode the
\emph{total} uncertainty in assigning a prediction to a given input.

Quantitatively, this reinterpretation can be achieved as follows.  We
invoke a normal approximation for the expected log-likelihood term in
Eqn. \ref{eqn:elbo_loss}, giving
\begin{equation}
    \label{eqn:normal_expected_logL}
    \mathbb{E}_{q(\mathbf{w} | \theta)} [\log p(\mathcal{D} | \mathbf{w})]
    = -\frac{1}{2} \sum_k \left[ \log{(2\pi)} + \log{(\sigma_k^2)}
    + \frac{(y_k - \mu_{y_k})^2}{\sigma_k^2} \right]
\end{equation}
where the variance $\sigma_k^2 \equiv \sigma_{a,k}^2 + \sigma_{e,k}^2$ is
the sum of the aleatoric variance $\sigma_{a,k}^2$ and the epistemic
variance $\sigma_{e,k}^2$, which we assume to be independent.  During training,
we optimize
the network parameters to minimize Eqn. \ref{eqn:elbo_loss} with Eqn.
\ref{eqn:normal_expected_logL} inserted appropriately, where $\sigma_k^2$
is now computed as the variance of the BNN output.  We
compute $\sigma_k^2$ using a moment propagation approach similar to Refs.
\cite{wu_deterministic_2019, hausmann_sampling-free_2020} which we describe
below, however $\sigma_k^2$ could also be computed as the BNN output
variance over repeated Monte Carlo samples of the network parameters.  With
this approach, the variance of a given BNN output will intrinsically
provide us with the sum of the aleatoric and epistemic variances,
without the need to separately predict the aleatoric variance.

\subsection{Simplified moment propagation}
\label{sec:moment_propagation}
Moment propagation allows us to analytically compute the mean and
variance of BNN outputs without the need for sampling
\cite{wu_deterministic_2019, hausmann_sampling-free_2020}, which is
desirable for lightweight BNN inference.  This is
done by propagating the mean and the variance sequentially through layers
of the BNN.  In other words, for each layer in the BNN, we take the mean
and variance of the input to that layer and then compute the mean and
variance of the corresponding output from that layer.  For many standard
network layers, such as convolutional layers, this can be done without
knowing the distribution of the input.  For more
complicated layers, we can often propagate the mean and variance
approximately by making simplifying assumptions (e.g., for a leaky-ReLU, we
can compute the mean and variance of the resulting output by assuming the
inputs are normally distributed).

In this section, we summarize the rules that we use to propagate
the mean and the variance through some common network layers.  We view our
approach to moment propagation as a somewhat simplified version of Ref.
\cite{wu_deterministic_2019} intended for use within the Bayes by Backprop
scheme \cite{blundell_weight_2015}, as we exclude the propagation of
covariance and incorporate a relatively simple, closed-form solution for
the mean and variance after propagation of a normal random variable through
a leaky-ReLU non-linearity.  We provide
additional details in Appendix \ref{appendix:moment_prop}.

\subsubsection{Fully connected layers}
For a fully connected layer with assumed independence between neurons, the mean
of the output $a_n'$
from the $n$-th Bayesian neuron is given by
\begin{equation}
    \label{eqn:fully_connected_mean}
    \mathbb{E}[a_n'] = \mathbb{E}[b_n] + \sum_i \mathbb{E}[a_i]
    \mathbb{E}[w_{n,i}]
\end{equation}
and the variance by
\begin{equation}
    \label{eqn:fully_connected_var}
    \mathbb{V}[a_n'] = \mathbb{V}[b_n] + \sum_i \mathbb{V}[a_i]
    \mathbb{V}[w_{n,i}]
    + \sum_i \mathbb{V}[a_i] \mathbb{E}[w_{n,i}]^2
    + \sum_i \mathbb{E}[a_i]^2 \mathbb{V}[w_{n,i}]
\end{equation}
where $b_n$ is the bias, $\mathbf{w}_n$ is the weight, and $\mathbf{a}$ is
the input from the previous layer.

\subsubsection{Convolutional layers}
\label{sec:moment_propagation_conv}
For a two-dimensional convolutional layer, we can simply rewrite Eqns.
\ref{eqn:fully_connected_mean} and \ref{eqn:fully_connected_var} in terms
of the appropriately sized weight, bias, and input images:
\begin{align*}
    \label{eqn:prop_convolutional}
    \mathbb{E}[\mathbf{a}'] &= \mathbb{E}[\mathbf{b}] + \mathbb{E}[\mathbf{a}]
    * \mathbb{E}[\mathbf{w}] \\
    \mathbb{V}[\mathbf{a}'] &= \mathbb{V}[\mathbf{b}] + \mathbb{V}[\mathbf{a}]
    * \mathbb{V}[\mathbf{w}]
    + \mathbb{V}[\mathbf{a}] * \mathbb{E}[\mathbf{w}]^2
    + \mathbb{E}[\mathbf{a}]^2 * \mathbb{V}[\mathbf{w}]
\end{align*}
where $*$ denotes the two-dimensional convolution operation as
defined in a given software implementation.  For example, the expected
value can be implemented by a ``conv2d'' operation as $\mathbb{E}[\mathbf{a}'] =
conv2d(\mathrm{input}=\mathbb{E}[\mathbf{a}],
\mathrm{weight}=\mathbb{E}[\mathbf{w}],
\mathrm{bias}=\mathbb{E}[\mathbf{b}])$.

\subsubsection{Average pooling layers}
Average pooling can be applied directly to the mean since $\mathbb{E}[X+Y] =
\mathbb{E}[X] + \mathbb{E}[Y]$ (assuming independence of neurons from one
another).  Similarly, average pooling can be applied
directly to
the variance with an additional scale factor of $1/N$ where $N$ is the
number of elements pooled by the average pooling operation.  The scale
factor arises as $\mathbb{V}[ \mathbb{E}[X]] = \mathbb{V}[(1/N) \sum_{i=1}^N
X_i] =
(1/N^2) \sum_{i=1}^N \mathbb{V}[X_i] = (1/N) \mathbb{E}[ \mathbb{V}[X] ]$.

\subsubsection{Leaky-ReLU}
To propagate the mean and variance through a leaky-ReLU, we assume that the
input to the leaky-ReLU is normally distributed.  Although this may not be
true in general, it appears reasonable for many network architectures with
normally distributed weights and biases \cite{wu_deterministic_2019}.
Assuming a normally distributed input $a_n$ to the leaky-ReLU,
we compute the mean from the law of total expectation as
\begin{equation}
    \mathbb{E}[a_n'] = l P(a_n<0) \mathbb{E}[a_n | a_n < 0] + P(a_n\geq0)
    \mathbb{E}[a_n | a_n \geq 0]
\end{equation}
and the variance from the law of total variance as
\begin{align}
    \mathbb{V}[a_n'] = &l^2 P(a_n<0) \mathbb{V}[a_n | a_n < 0]
    + P(a_n\geq0) \mathbb{V}[a_n | a_n \geq 0] \nonumber \\
    &+ l^2 \mathbb{E}[a_n | a_n < 0]^2 P(a_n<0) P(a_n\geq0) \nonumber \\
    &+ \mathbb{E}[a_n | a_n \geq 0]^2 P(a_n\geq0) P(a_n<0) \nonumber \\
    &- 2 l \mathbb{E}[a_n | a_n < 0] \mathbb{E}[a_n | a_n \geq 0] P(a_n<0)
    P(a_n\geq0)
\end{align}
where $l$ is the slope of the leaky-ReLU for arguments below 0; $P(a_n<0)$
and $P(a_n \geq 0)$ are the probabilities of $a_n$ being below 0 or above
0, respectively, assuming it is normally distributed; $\mathbb{V}[a_n | a_n <
0]$
and $\mathbb{V}[a_n | a_n \geq 0]$ are the variances of $a_n$ assuming a
truncated normal distribution truncated below 0 or above 0, respectively;
and $\mathbb{E}[a_n | a_n < 0]$ and $\mathbb{E}[a_n | a_n \geq 0]$ are the
means of $a_n$
assuming a truncated normal distribution truncated below 0 or above
0, respectively.  Each of these terms are written out explicitly below:
\begin{align*}
    \Phi(\alpha) \equiv P(a_n<0) &= \frac{1}{2} \left[1 + \mathrm{erf}
    \left( -\frac{\alpha}{\sqrt{2}} \right)
    \right] \\
    P(a_n \geq 0) &= 1 - P(a_n<0) \\
    \mathbb{V}[a_n | a_n < 0] &= \mathbb{V}[a_n] \left\{1 - \alpha
    \frac{\phi(\alpha)}{\Phi(\alpha)}
    - \left[ \frac{\phi(\alpha)}{\Phi(\alpha)} \right]^2
    \right\} \\
    \mathbb{V}[a_n | a_n \geq 0] &= \mathbb{V}[a_n] \left\{1 + \alpha
    \frac{\phi(\alpha)}{1-\Phi(\alpha)}
    - \left[ \frac{\phi(\alpha)}{1-\Phi(\alpha)} \right]^2
    \right\} \\
    \mathbb{E}[a_n | a_n < 0] &= \mathbb{E}[a_n] - \sqrt{\mathbb{V}[a_n]}
    \frac{\phi(\alpha)}{\Phi(\alpha)} \\
    \mathbb{E}[a_n | a_n \geq 0] &= \mathbb{E}[a_n] + \sqrt{\mathbb{V}[a_n]}
    \frac{\phi(\alpha)}{1-\Phi(\alpha)}
\end{align*}
where we have defined $\alpha=-\mathbb{E}[a_n]/\sqrt{\mathbb{V}[a_n]}$ and
$\phi(\alpha)=\exp(-\alpha^2 / 2) / \sqrt{2\pi}$.  To avoid numerical
issues resulting from very small or very large values of $\alpha$
(corresponding to normal distributions that nearly entirely lie below or
above 0, respectively), we clamp $\alpha$ to ensure $|\alpha| \leq 3$.

\section{Experimental demonstration}
\begin{figure}[htbp!]
    \centering
    \includegraphics[]{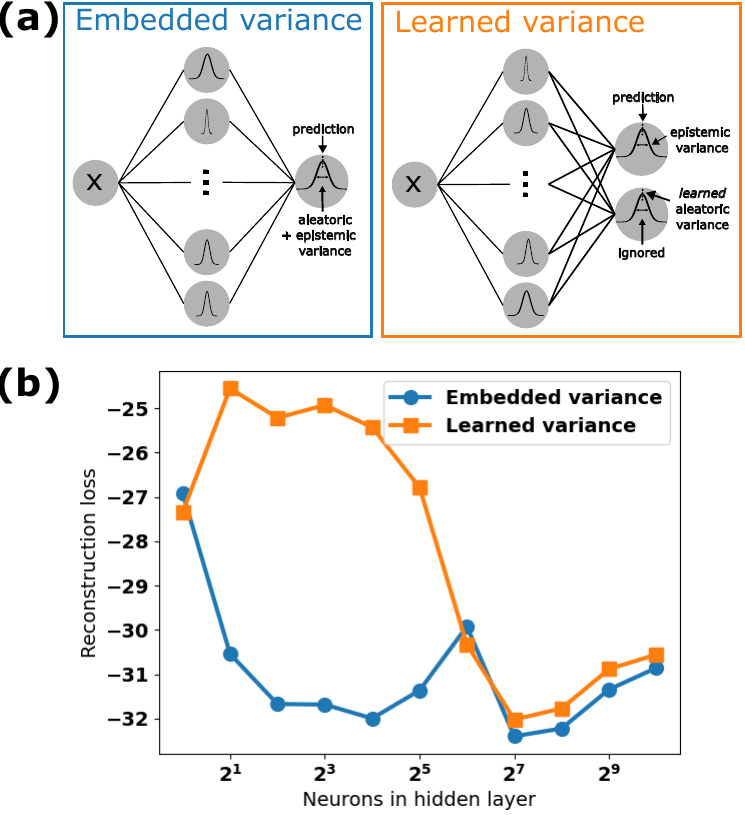}
    \caption{\textbf{BNN architecture comparison and performance.} (a)
        Architecture diagrams showing the BNN architecture used with the
        ``\methodname{}'' approach (left) versus the ``\comparename{}''
        approach (right) for recovering heteroscedastic variance.  Each BNN
        contains a single hidden layer with leaky-ReLU activation functions.
        Notably, for an equal number of neurons in the hidden layer, the
        ``\comparename{}'' consists of a greater number of learnable
        parameters.
        (b) Reconstruction loss (expected negative log-likelihood) versus
        the
        number
        of neurons in the hidden layer of each BNN.  Results for the BNN
        trained with the ``\methodname{}'' approach (shown in (a) on the
        left)
        are indicated by blue filled circles.  Results for the BNN trained
        with
        the ``\comparename{}'' approach (shown in (a) on the right) are
        indicated by orange filled squares.}
    \label{fig:poly_data_loss}
\end{figure}

To demonstrate the performance of our proposed framework for variational
inference of lightweight BNNs, we compare the results of our method to
those obtained
when learning the aleatoric variance as an additional network output.  For
consistency of comparison, all comparisons are made using the same moment
propagation technique described in Section \ref{sec:moment_propagation},
with the only difference being the splitting of the output head needed when
learning the aleatoric variance as a network output as in
\cite{nix_1994, an_variational_2015, kendall_what_2017, wu_deterministic_2019,
    antoran_getting_2021} (see
Fig. \ref{fig:poly_data_loss}(a)).  We
refer to our method (i.e., embedding the combined aleatoric and epistemic
uncertainty into the variances of BNN parameters) as ``\methodname{}''.  We
refer to the split output head approach (i.e., learning the aleatoric
variance as an
additional output of the BNN) as ``\comparename{}''.

We compare the ``\methodname{}'' approach to the ``\comparename{}''
approach by solving a polynomial regression problem with a simple BNN
consisting of fully connected layers with leaky-ReLU activation functions,
where data consists of a noisy polynomial generated by a heteroscedastic
noisy process.  We train both networks on the same training data and then
compare their predictions on unobserved testing data, with test data that
is both similar to and dissimilar to the training data.  We describe this
experiment and the results in the following sections.


\subsection{Polynomial regression}
\label{sec:poly_exp_description}
To demonstrate that our proposed lightweight inference framework is capable
of recovering heteroscedastic aleatoric and epistemic uncertainties, we
apply it to a simple polynomial regression problem with heteroscedastic
noise.  Specifically, we use a multi-layer fully connected BNN with
leaky-ReLU activation functions to predict data generated by
\begin{align}
    \label{eqn:poly_model}
    &y = x + \epsilon(x) + 1 \\
    &\epsilon(x) \sim \mathcal{N}(0, \sigma(x)^2) \nonumber
\end{align}
where $\epsilon(x)$ is heteroscedastic aleatoric Gaussian measurement noise
with mean $\mu=0$ and variance $\sigma(x)^2 = \left[ 0.1 + 0.2 \cdot
\sin{(2 \pi
    x -
    \pi/2)} \right]^2$.  For this experiment, we use a single hidden layer and
vary the number of Bayesian neurons in the hidden layer (see Fig.
\ref{fig:poly_data_loss}(a)).  We then inspect
the reconstruction loss and qualitative plots of predictions with
uncertainties for both the ``\methodname{}'' and the ``\comparename{}''
methods.

\subsubsection{Training details}
To compare the ``\methodname{}'' and ``\comparename{}'' methods for
inference, we trained the networks described above for 10000 epochs,
generating 64 fresh samples of
training data shared by each network at each epoch.  This was done by
sampling $x$ from a uniform distribution over the support of the training
distribution ($x \in [-0.5, 0.5]$) and then
generating the ground truth $y$ according to Eqn. \ref{eqn:poly_model}.
Since the
BNNs never
saw the same data twice, we adapted the KL decay scheme from Ref.
\cite{blundell_weight_2015} so that the first term in Eqn.
\ref{eqn:elbo_loss}
decays after each epoch $i=1,2,\ldots,M$ by a multiplicative factor
$\alpha(i) = \frac{2^{M-i}}{2^M-1}$, which we view as an analogous (but
simpler) procedure to updating the prior distribution after each new
(batched) data
observation.  We used a single spike slab prior common to all weights and
biases defined as $p(\cdot) = 0.5 \left[ N(0, \sigma_1^2) + N(0,
\sigma_2^2)\right]$ with $\sigma_1^2 = 1$ and $\sigma_2^2 = e^{-12}$.  We
approximated the KL divergence by Monte Carlo sampling as in Eqn.
\ref{eqn:kl_monte_carlo} with a single sample ($N=1$).  We used the AdamW
\cite{loshchilov_decoupled_2019} optimizer
with a learning rate $\alpha=0.01$, $\beta_1=0.9$, $\beta_2=0.999$,
$\epsilon=10^{-8}$, and a
weight decay $\lambda=0.01$.  Presented results
(e.g., figures of
predictions) were generated from the models with the minimum reconstruction
losses on fresh validation data across all training epochs.

\begin{figure}[htbp!]
    \centering
    \includegraphics[width=\textwidth]{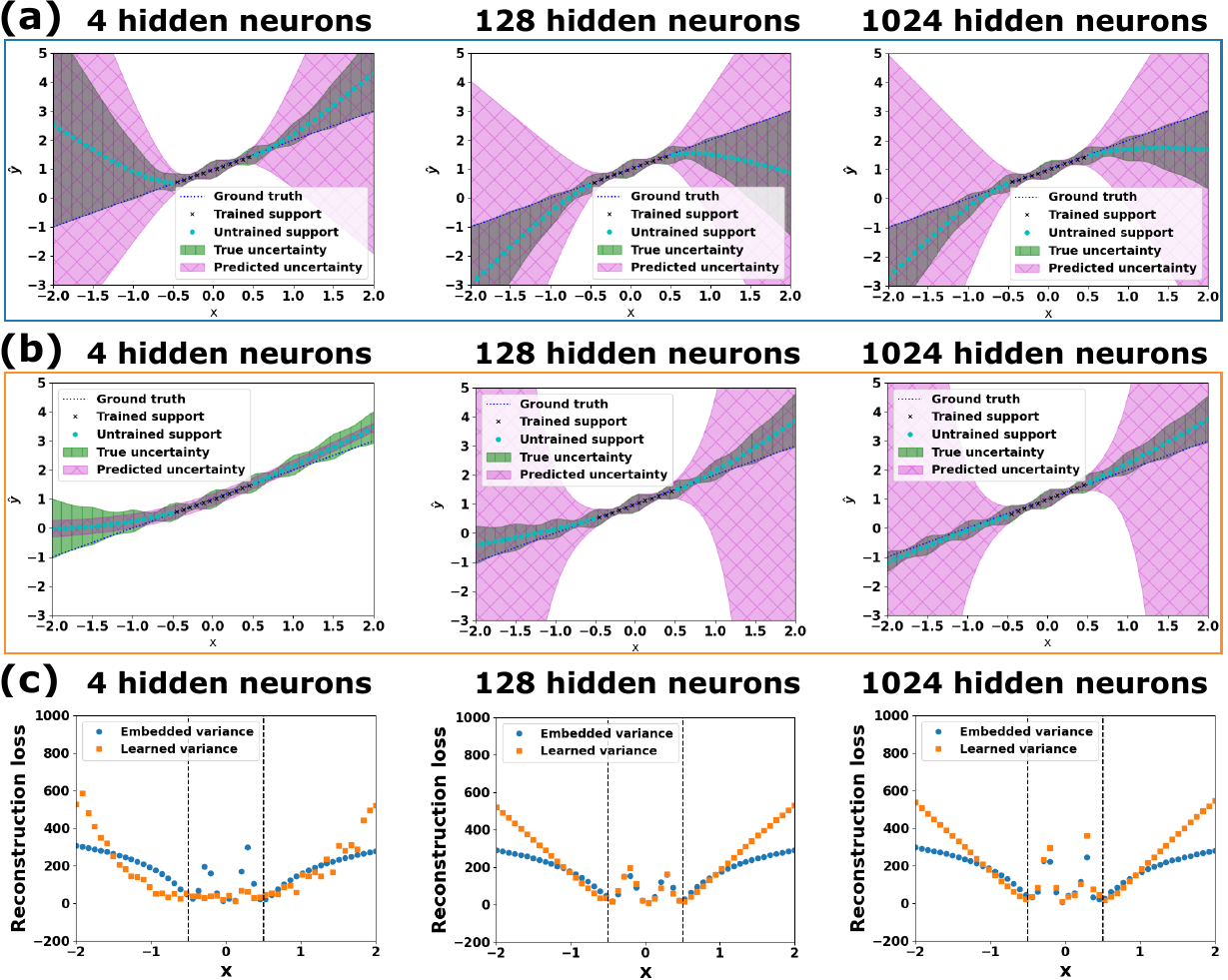}
    \caption{\textbf{Heteroscedastic variance recovered by BNN.} BNN
        predictions with uncertainties when predicting the data $y = x +
        \epsilon(x) + 1$ with heteroscedastic noise $\epsilon(x)$.  (a)
        Predictions from the BNNs
        trained with the ``\methodname{}'' approach (this work), where the
        combined aleatoric and
        epistemic variances are embedded into the variances of the
        trainable parameters of the BNN.  (b) Predictions from the BNNs
        trained with the ``\comparename{}'' approach, where the aleatoric
        variance is predicted as an output of the BNN.  (c) Reconstruction
        losses on data sampled from the support of the training
        distribution (in the range $[-0.5, 0.5]$) and data sampled from
        outside of this support.  The dashed black lines indicate the
        boundaries of the training distribution range $[-0.5, 0.5]$.  The
        reconstruction losses indicated by blue circles correspond to the
        models shown in (a), while the losses indicated by orange squares
        correspond to the models shown in (b).}
    \label{fig:poly_data_qualitative}
\end{figure}

\subsubsection{Results and discussion}
Predictions made by the trained BNNs are shown in Fig.
\ref{fig:poly_data_qualitative},
where fresh testing data were spaced evenly both within the
range of the training data ($x \in [-0.5, 0.5]$) to observe recovery of
aleatoric uncertainty and out of the range of training data to
observe recovery of epistemic uncertainty.  With only 4 neurons in the
hidden layer of each BNN, the ``\methodname{}'' model accurately predicts
the ground truth trend and uncertainty for in-distribution data, while
expressing high uncertainty for out-of-distribution data (Fig.
\ref{fig:poly_data_qualitative}(a)).  In contrast,``\comparename{}'' BNN
fails to recover the true uncertainty for in-distribution data, and greatly
underestimates its uncertainty on out-of-distribution data (Fig.
\ref{fig:poly_data_qualitative}(b)).  With 128 hidden
neurons, both models appear to perform similarly qualitatively (Fig.
\ref{fig:poly_data_qualitative}(a,b)), however the ``\methodname{}'' model
appears to generalize more favorably to out-of-distribution data in terms
of the reconstruction loss (Fig. \ref{fig:poly_data_qualitative}(c)).  With
a very large number (e.g., 1024) of neurons in the hidden layer, both inference
approaches appear to perform similarly.  Notably, by keeping the number of
neurons
in the hidden layer fixed, the ``\comparename{}'' BNN has more learnable
parameters than the ``\methodname{}'' BNN.  For instance, just before the
crossover point in the reconstruction loss at 128 hidden neurons in Fig.
\ref{fig:poly_data_loss}(b), the ``\comparename{}'' BNN has about 34\% more
learnable parameters than the ``\methodname{}'' BNN (1028 versus 770).

The
minimum reconstruction loss (expected negative log-likelihood) versus the
number of
hidden neurons for each model is shown in Fig.
\ref{fig:poly_data_loss}(b).  For lightweight BNNs, our
``\methodname{}'' approach achieves improved performance in
terms of the reconstruction loss.  For heavier BNNs with a greater number
of learnable parameters, the ``\methodname{}'' and ``\comparename{}''
models appear to perform similarly, with ``\comparename{}'' models
sometimes outperforming our ``\methodname{}'' models on data similar to the
training data.  These results suggest that both inference approaches be
considered when designing a model, as their respective regimes of
superiority may vary with desired model size.

Interestingly, the ``\methodname{}'' approach seems to generalize more
favorably to out-of-distribution data.  Specifically, comparing the
reconstruction losses for the models shown in Figs.
\ref{fig:poly_data_qualitative}(a,b), the losses for the ``\methodname{}''
models appear to increase more gradually than those for the
``\comparename{}'' models of a similar size (see Fig.
\ref{fig:poly_data_qualitative}(c)).  We believe that a deeper
investigation of this behavior may be fruitful.


%
%

\section{Conclusions}
In this work, we demonstrated that learning aleatoric predictive
uncertainties as an additional output of a BNN can reduce predictive
performance,
requiring a greater number of learnable parameters to be incorporated in
the model architecture.
We proposed a reinterpretation of the variances of
learnable BNN parameters which allows us to extract both the aleatoric and
epistemic uncertainties directly from the variances accumulated on a
forward pass through the BNN, without the need for learning aleatoric
uncertainties as an additional output.  In doing so, our reinterpretation
effectively embeds the \emph{total} predictive uncertainty into the already
present learnable parameters of a BNN, enabling meaningful heteroscedastic
uncertainty predictions from comparatively lightweight BNNs.  By combining
this reinterpreted approach to extracting uncertainties from a BNN with a
simplified moment propagation technique for BNN inference, we have
developed a simple, sampling-free, and performant framework suitable for
application to lightweight BNNs.

\begin{ack}
    This material is based partially upon work supported by the Defense
    Advanced Research Projects Agency (DARPA) under Contract No.
    HR001120C0027.  The
    views, opinions and/or findings expressed are those of the author and should
    not be interpreted as representing the official views or policies of the
    Department of Defense or the U.S. Government.  Approved for Public Release,
    Distribution Unlimited.
\end{ack}

\clearpage{}
\bibliographystyle{unsrt}
\bibliography{references/ref_file_primary.bib}

\appendix
\section{Moment propagation through layers of neurons}
\label{appendix:moment_prop}
Focusing on a single fully connected layer, we write the output $a_n'$ of
the $n$-th Bayesian neuron in the layer as
\begin{equation}
    \label{eqn:fc_layer_output}
    a_n' = b_n + \sum_j w_{n,j} a_j
\end{equation}
where $b_n$ is a bias distributed with mean $\mu_{b_n}$ and variance
$\mu_{b_n}$, $w_{n,j}$ are weights distributed with mean $\mu_{n,j}$
and variance $\sigma_{n,j}^2$, and the inputs $a_j$ from the previous layer
are distributed with mean $\mu_j$ and variance $\sigma_j^2$.  For two
independent random variables $X$ and $Y$, we note the following properties:
\begin{align}
    \mathbb{E}[XY] &= \mathbb{E}[X] \mathbb{E}[Y] \label{eqn:expected_value} \\
    \mathbb{V}[XY] &= \mathbb{V}[X] \mathbb{V}[Y] + \mathbb{V}[X]
    \mathbb{E}[Y]^2 +
    \mathbb{E}[X]^2 \mathbb{V}[Y] \label{eqn:variance}
\end{align}
Assuming that the inputs $a_j$ are conditionally independent from
parameters $b_n$ and $w_{n,j}$ given the data $\mathcal{D}$, we can apply
Eqns. \ref{eqn:expected_value} and \ref{eqn:variance} (as well as other
common properties of expectation and variance) to Eqn.
\ref{eqn:fc_layer_output} to obtain
\begin{align*}
    \mathbb{E}[a_n'] &= \mathbb{E}[b_n] + \sum_j \mathbb{E}[w_{n,j}]
    \mathbb{E}[a_j] \\
    \mathbb{V}[a_n'] &= \mathbb{V}[b_n] + \sum_i \mathbb{V}[a_i]
    \mathbb{V}[w_{n,i}]
    + \sum_i \mathbb{V}[a_i] \mathbb{E}[w_{n,i}]^2
    + \sum_i \mathbb{E}[a_i]^2 \mathbb{V}[w_{n,i}]
\end{align*}
This result is readily extended to convolutional layers as described in
Section \ref{sec:moment_propagation_conv}.

\end{document}